\pdfoutput=1

\documentclass[11pt]{article}

\usepackage{naacl2020}

\usepackage[russian,english]{babel}
\usepackage{flushend}
\usepackage{times}
\usepackage{latexsym}
\usepackage{comment}
\usepackage{amsmath,bm}
\usepackage[T1]{fontenc}

\usepackage[utf8]{inputenc}

\usepackage{graphicx}
\usepackage{microtype}

\title{Gender Bias in Multilingual Neural Machine Translation: \\The Architecture Matters}

\author{Marta R. Costa-jussà, Carlos Escolano, Christine Basta, Javier Ferrando, \\\textbf{Roser Batlle and Ksenia Kharitonova} \\
TALP Research Center, Universitat Polit\`ecnica de Catalunya, Barcelona\\
  \texttt{\{marta.ruiz,carlos.escolano,christine.raouf.saad.basta\}@upc.edu}\\\texttt{javier.ferrando.monsonis@upc.edu,roser.batlle@estudiantat.upc.edu}\\\texttt{ksenia.kharitonova@estudiantat.upc.edu}}

\begin{document}
\maketitle
\begin{abstract}
Multilingual Neural Machine Translation architectures mainly differ in the amount of sharing modules and parameters among languages. In this paper, and from an algorithmic perspective, we explore if the chosen architecture, when trained with the same data, influences the gender bias accuracy. Experiments in four language pairs show that Language-Specific encoders-decoders exhibit less bias than the Shared encoder-decoder architecture. Further interpretability analysis of source embeddings and the attention shows that, in the Language-Specific case, the embeddings encode more gender information, and its attention is more diverted. Both behaviors help in mitigating gender bias.

\end{abstract}

\section{Introduction}

Machine translation has been shown to exhibit gender bias \cite{prates:2018}, and several solutions have already been proposed to mitigate it \cite{johnson,font:2019,costajussa:2020}. The general gender bias in Natural Language Processing (NLP) has been mainly attributed to data \cite{costajussa:2019}. Several studies show the pervasiveness of stereotypes in book collections \cite{madaan:2018}, or Bollywood films \cite{bollywood}, among many others. As a consequence, our systems trained on this data exhibit biases. Among other strategies, several studies have proposed to work in data augmentation to balance data \cite{zmigrod-etal-2019-counterfactual} or forcing gender-balanced datasets \cite{kelliegap:2018, costa-jussa-etal-2020-gebiotoolkit}. In parallel, other initiatives focus on documenting our datasets \cite{gebru} to prioritize transparency.

However, data is not the only reason for biases, and recent studies show that 
our models can be trained in a robust way to reduce the effects of data correlations (e.g., stereotypes, among others). In \cite{webster2020measuring}, the authors explored available mitigations and by increasing dropout, which resulted in improving how the models reasoned about different stereotypes in WinoGender examples \cite{rudinger:2018}. 

The purpose of the current paper is to explore if the Multilingual Neural Machine Translation (MNMT) architecture can impact the amount of gender bias. To answer this question, we compare MNMT architectures trained with the same data and quantify their amount of gender bias with the standard WinoMT evaluation benchmark \cite{stanovsky-etal-2019-evaluating}. Results show that the Language-Specific encoders-decoders \cite{escolano:2020} exhibit less bias than the Shared encoder-decoder \cite{johnson2017google}. Then, we analyze and visualize why the MNMT architecture impacts mitigating or amplifying this bias by studying its internal workings. We study the amount of gender information that the source embeddings encode, and we see that Language-Specific surpasses Shared in these terms, allowing for a better prediction of gender.

Additionally, and taking advantage that both Shared and Language-Specific are based on the Transformer \cite{vaswani_attention:2017}, we study the coefficient of variation in the attention \cite{kobayashi-etal-2020-attention}, which shows that the attention span is narrower for the Shared system than for the Language-Specific one. Therefore, the context taken into account is smaller for the Shared system, which causes a higher gender bias. 


Finally, we also do a manual analysis to investigate which biases have a linguistic explanation. 

\section{Background: Multilingual Architectures}
\label{sec:arch}

In this section, we briefly describe the MNMT architectures that we are exploring. While most NMT architectures are based on the encoder-decoder architecture, behind this architecture we can have either LSTMs \cite{Sutskever:2014}, Transfomers \cite{vaswani_attention:2017} or a mixture of both \cite{chen-etal-2018-best}. The NMT is then extended to many-to-many multilingual languages using a shared encoder-decoder, or having one encoder and decoder for each language. 

\paragraph{Transformer.}
In the encoder-decoder approach, the source sentence is encoded into hidden state vectors, whereas the decoder predicts the target sentence using the last representation of the encoder.
The Transformer utilizes multi-head attention in different manners: encoder self-attention, decoder self-attention, and decoder-encoder attention. Positional embeddings are applied to both the encoder and decoder. This substitutes the recurrent operations in LSTM, and no sequential processing is needed. 

\paragraph{Shared Encoder-Decoder.} \citet{johnson2017google} train a single encoder and decoder with multiple input and output languages. Given a language set, a shared architecture has a universal encoder and decoder fed with all initial language pairs at once. The model shares vocabulary and parameters among languages to ensure that no additional ambiguity is introduced in the representation. By sharing a single model across all languages, the system can represent all languages in a single space. This allows translation between language pairs never seen during the training process, which is known as zero-shot translation.

\paragraph{Language-Specific Encoders-Decoders.} Architectures in this category may vary between sharing some layers \cite{firat:2017,lu-etal-2018-neural} or no sharing at all \cite{escolano:2020}. This paper uses the latter since it is the most contrastive to the shared encoder-decoder. The Language-Specific (with no sharing) approach consists of training independent encoders and decoders for each language. Different from the standard pairwise training is that, in this case, there is only one encoder and one decoder for each language. Since parameters are not shared, this joint training enables new languages without the need to retrain the existing modules, which is a clear advantage to previously shared encoder-decoder.

\section{Experimental Framework}
In this section, we report the details of the experiments that we are running.

\subsection{Data and Parameters}

Experiments are performed on the EuroParl data \cite{koehn:2005} on English, German, Spanish and French, with parallel sentences among all combinations of these four languages, with approximately 2 million sentences per language pair. For English-Russian, we used 1 million training sentences from the \textit{Yandex} corpus \footnote{\url{https://translate.yandex.ru/corpus?lang=en}}. We build two sets of MNMT systems: excluding or including Russian. We do our experiments with two sets to measure how these models react to a new language that is not related to any of the previous ones written using a different set of characters. When excluding Russian, systems are trained with English, German, Spanish, and French with parallel sentences among all four languages. When including Russian, systems are trained with previous data plus the English-Russian pair. As contrastive systems, we also built the pairwise Bilingual systems (based on the Transformer) on the corresponding language pair data. As validation and test set, we used \textit{newstest2012} and \textit{newstest2013} from WMT\footnote{\url{http://www.statmt.org}}.
All data were preprocessed using standard Moses scripts \cite{koehn2007moses}.

Experiments were done using the implementation provided by Fairseq\footnote{Release v0.6.0 available at \url{https://github.com/pytorch/fairseq}}.
We used 6 layers, each with 8 attention heads,
an embedding size of 512 dimensions,
and a vocabulary size of 32k subword tokens with 
Byte Pair Encoding \cite{sennrich-etal-2016-neural} (per pair). Dropout was 0.3 and trained with an effective batch size of 32k tokens for approximately 200k updates, using the validation loss for early stopping. In all cases, we used Adam \cite{kingma2014adam} as the optimizer, with a learning rate of 0.001 and 4000 warmup steps. 
All experiments were performed on an NVIDIA Titan X GPU with 12 GB of memory. %

\subsection{WinoMT: Gender Bias Evaluation}
\label{sec:winoMT}

WinoMT \cite{stanovsky-etal-2019-evaluating} is the first challenge test set for evaluating gender bias in MT systems. This test set consists of 3888 sentences. On the one side, the test set is distributed with 1826 male sentences, 1822 female sentences and 240 neutral sentences. On the other side, the test set is distributed with 1584 anti-stereotyped sentences, 1584 pro-stereotyped sentences, and 720 neutral sentences. 
Each sentence contains two personal entities, where one of them is a co-referent to a pronoun and a golden gender is specified for this entity.
An example of the anti-stereotyped sentences is: 

\begin{quote}
    'The \textit{developer} argued with the \textit{designer} because \textit{she} did not like the design.'
\end{quote}

\noindent where \textit{she} refers to the \textit{developer}. \textit{developer} is considered the golden entity with feminine as the gender.  The same sentence would be pro-stereotyped if \textit{she} was replaced with \textit{he}, referring to the developer as masculine word.

The evaluation depends on comparing the translated entity with the specified gender of the golden entity to correctly gendered translation. 
Three metrics were used for assessment: accuracy (Acc.), \bm{$\Delta G$} and \bm{$\Delta S$}. The accuracy is the correctly inflected entities compared to their original golden gender. \bm{$\Delta G$} is the difference between the correctly inflected masculine and feminine entities. \bm{$\Delta S$} is the difference between the inflected genders of the pro-stereotyped entities and the anti-stereotyped entities. We report the three parameters for each system for the four languages, German, Spanish, French, and Russian. These four languages are gendered concerning the English, as they have gender mark for the nouns, adjectives, and articles. Consequently, we examine which languages are more challenging to have correct translations when translating from a non-gendered language like English.

\begin{table*}[h]
    \centering
    \begin{tabular}{|c|l||c|c|c|c||c|c|c|c|}
    \hline
        \multicolumn{2}{|c||}{Language Set}& \multicolumn{4}{c||}{en,de,es,fr}& \multicolumn{4}{c|}{en,de,es,fr,ru}\\\hline
        Language &System & BLEU$\uparrow$ & Acc$\uparrow$ & \bm{$\Delta G$}$\downarrow$  & $\Delta S$$\downarrow$ & BLEU$\uparrow$ & Acc$\uparrow$ & \bm{$\Delta G$}$\downarrow$  & $\Delta S$$\downarrow$ \\ \hline
         \hline
         ende  &bil & 21.61 &\bf 64.10 & \bf 5.7 & 8.30  & 21.61 & 64.10 & \bf 5.7 & 8.30 \\ \cline{2-10}
         & shared &21.39 &53.86 &23.59 &8.33 & \bf 22.11 & 51.65 & 28.48 & \bf 6.82\\ \cline{2-10}
         & lang-spec   & \bf 22.01& 56.28 &  17.45& \bf 7.83 & 22.06 & \bf 65.48 & 8.62& 8.09\\ \hline
         enes &bil & 25.82 &46.00 &  22.90 & \bf 2.40 & 25.82 &46.00 &  22.90 & \bf 2.40\\ \cline{2-10}
         & shared &28.08 &51.67 &24.77 &5.49 & 29.78 & 44.78 & 43.35 &6.82\\ \cline{2-10}
         & lang-spec  & \bf 29.53& \bf 54.19 & \bf 20.73&7.64 & \bf 29.97 & \bf 57.90 & \bf 14.60 & 21.47 \\ \hline
         enfr &bil & 26.73 &42.18 & \bf 21.59 & 14.16 & 26.73 &42.18 & 21.59 & 14.16\\ \cline{2-10}
         & shared &28.43 & 45.55 & 24.99& \bf0.06 & 29.63 & 43.67 & 36.86 & \bf2.34\\ \cline{2-10}
         & lang-spec  & \bf 29.74 & \bf 45.81&28.45&5.64 & \bf 29.92  & \bf 49.10 & \bf 21.51 & 11.43\\ \hline
         enru &bil & \multicolumn{4}{c||}{-}&  22.98 & \bf 35.50& 36.70& \bf 4.1  \\ \cline{2-10}
         & shared & \multicolumn{4}{c||}{-} & 20.3 & 35.44 & \bf 33.00& 6.32 \\ \cline{2-10}
         & lang-spec  & \multicolumn{4}{c||}{-} & \bf 23.94 & 32.07 & 36.01& 11.62\\ \hline

    \end{tabular}
    \caption{Results in terms of BLEU and Gender Accuracy for the 3 compared systems: Bilingual (bil), Shared (shared) and Language-Specific (lang-spec) and two language sets (without and with Russian).}
    \label{tab:lstmtrans}
\end{table*}

\section{Results}

Table \ref{tab:lstmtrans} reports the results in terms of BLEU and gender accuracy for the architectures described in section \ref{sec:arch}. This table reports results in two different language sets: without or with Russian.

\paragraph{Bilingual vs Multilingual.} Coherently with previous studies \cite{johnson2017google}, multilingual systems improve bilingual systems in terms of translation quality. However, and depending on the multilingual architecture that we are using, we can not conclude the same in terms of gender accuracy. For the case study without Russian, the Bilingual architecture improves the Shared one in two out of the three language pairs, only in one out of the three language pairs the multilingual Language-Specific architecture. For the case study with Russian, the Language-Specific architecture improves in all pairs, except for Russian. Regarding \bm{$\Delta G$}, the Bilingual system tends to be better only for the language set without Russian. This does not hold for the language set with Russian, where Language-Specific is better. Finally, regarding \bm{$\Delta S$} we observe that best systems tend to be Bilingual or Shared depending on the language pair.



\paragraph{Shared vs Language-Specific.} When comparing these two architectures, we observe that the Language-Specific architecture shows consistent gains in terms of BLEU, around 0.2-3.6\%. The only exception where the Shared system outperforms the Language-Specific one is English-German (trained with Russian). Such superiority of the Language-Specific system becomes higher in terms of gender accuracy, where improvements surpass 13\% for German and Spanish (trained with Russian). Differences are smaller in training without Russian, but still better for the Language-Specific architecture for all languages. 

When comparing \bm{$\Delta G$}, conclusions are similar, with the Language-Specific system showing gains up to 29\%. Note that since WinoMT is divided into 46,97\% male, 46,86\% female and 6.17\% neutral (see section \ref{sec:winoMT}), 46\% accuracy can be easily achieved by predicting most of the time the same gender.
For the Shared architecture, we observe that the high \bm{$\Delta G$} is explained by having a strong preference for predicting male gender.
In both cases, the big exception is gender bias accuracy for Russian, where the Shared system slightly improves over the Language-Specific system. Still, in this case, performance for all systems is close to random. 

Finally, regarding \bm{$\Delta S$}, results tend to be better for the Shared architecture. These differences in \bm{$\Delta S$} are because accuracy in Shared, for both pro and anti-stereotypical occupations, is much lower than for Language-Specific, which derives from having lower differences. Overall, we can conclude that gender accuracy is much better for Language-Specific.

We observe that while training with the same data, the gender bias accuracy that we get is significantly different when using different architectures, with clear outperformance of the Language-Specific over the Shared one. To bring some light to these results, we are providing a deep interpretability and manual analysis in the following sections. 

\section{Interpretability of the Results: Source Embeddings and Attention Analysis}
\label{sec:vis}

This section aims to measure gender bias on different multilingual architectures to understand their difference when encoding source sentences and decoding target ones. We will focus on the two main components of the Transformer architecture that can be the origin of such biases, the contextual representation of the source tokens created by the encoder, and the encoder-decoder attention. We perform the same experiments on Shared and Language-Specific architectures, trained on the same data, and Bilingual systems, trained on the same data using only one of the language pairs. 



\subsection{Source Embeddings}
\label{sec:svm}

\paragraph{Goal.} We want to study how source contextual embeddings codify gender information. Previous works \cite{basta:2019} used a classification approach to verify that contextual embeddings contain gender information in English neutral occupations. While in \cite{basta:2019} means that contextual embeddings are biased, in our case, encoding information of gender in the source embeddings will help to predict the gender appropriately.

\paragraph{Classification Approach.} 
We choose two word-types for this classification using the provided information by WinoMT (see an example of a sentence in section \ref{sec:winoMT}): determiners (\textit{The}) and occupations, to measure how gender information is reflected at their contextual embeddings. The first category is initially neutral as it is equally employed in all categories. Therefore all gender information present in these embeddings must come from the context of the sentence. For each system and word-type, we train an SVM \cite{cortes1995support} classifier with Radial Basis Function kernel on 1000 randomly selected sentences from WinoMT and tested on the remaining 2888 sentences from the set. Words are represented as their first subword in case they are split in the vocabulary.
We performed 10 independent experiments to guarantee the randomization of token representations. Obtaining a higher accuracy in the classification results means that more information on gender is encoded in the source embeddings. 

\begin{figure*}[h!]
    \centering
    \includegraphics[width=0.95\textwidth]{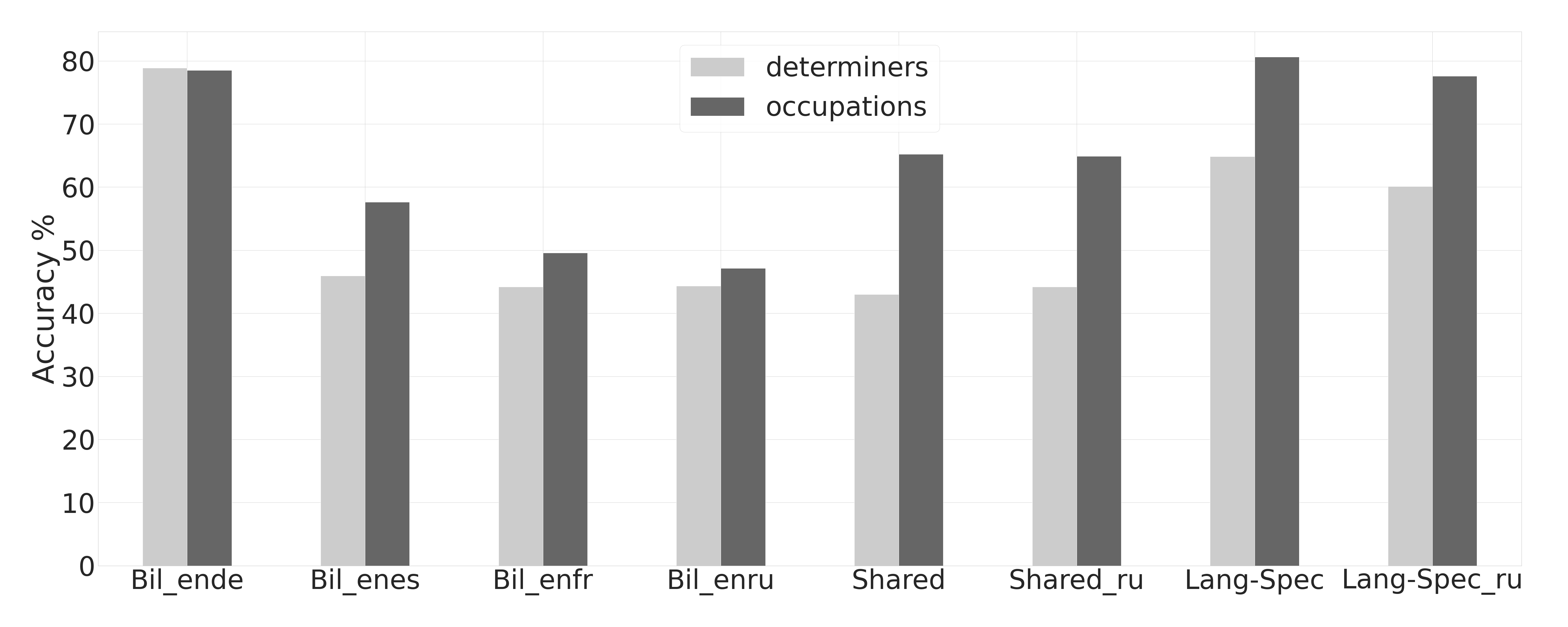}
    \caption{Classification Results, in light grey we have the determiner results and in dark grey the occupations results.}
    \label{fig:SVM}
\end{figure*}

\paragraph{Results.} Figure \ref{fig:SVM} shows the results for this classification for all the bilingual and multilingual systems (from left-to-right) for both determiners and occupations. Similarly to the previous section, we compare multilingual architecture in excluding Russian (shared, lang-spec) or including it (shared$\_$ru, lang-spec$\_$ru).

\paragraph{Bilingual Comparison.} Bilingual systems show that target language substantially impacts the amount of gender information encoded in the contextual representations. While the translation results are similar between all language pairs, the English-German outperforms by a significant margin (30\%) all other pairs even when trained on the same domain and using similar training set sizes. These results correlate with the gender accuracy in Table \ref{tab:lstmtrans} showing that the systems that encode more gender information on their contextual representations produce more gender accurate translations. 

\paragraph{Shared vs. Language-Specific.} When comparing multilingual systems, we see that the Language-Specific approach outperforms the Shared method on both determiners and occupations, showing to include more gender information. In contrast, the Shared architecture focuses more on the occupations, which is associated with stereotypes. 
In this case, we observe the same correlation between contextual representation and gender accuracy in translation, being the Language-Specific architecture only outperformed by the Bilingual English-German.

\paragraph{Languages Set (without or with Russian).} The amount and type of languages that are used to train MNMT models have a shallow impact on the amount of gender information encoded by the contextual vectors. When adding Russian, experiments show that both Shared and Language-Specific show a slight variation (from 1\% gain to 3\% loss). 


\subsection{Attention}

\paragraph{Goal.} We want to study the degree of softness of the distribution of the attention scores to possibly interpret the results in gender accuracy from Table \ref{tab:lstmtrans}. For this, we use attention vector norms and a coefficient of variation to measure how much diverted the attention is in each system. 

\paragraph{Attention Vector Norms.} Since both Shared and Language-specific models use the Transformer, we can compare their behaviors using the same interpretability method. Previous works have studied the relationship between the attention mechanism and the model output \cite{clark-etal-2019-bert,kovaleva-etal-2019-revealing,NEURIPS2019_159c1ffe,lin-etal-2019-open,marecek-rosa-2019-balustrades,Htut2019DoAH,raganato-tiedemann-2018-analysis,tang-etal-2018-analysis}.

The encoder-decoder attention mechanism computes a vector representation $attn_t$ based on the source sequence's encoder output representations $encoder\_out$, as keys and values, and the previous module output $attn_{t}*$ as query. These inputs are linearly transformed by matrices $W^K,W^V$ and $W^Q$ respectively, getting $K, V$ and $q_{t}$. Attention weights $\alpha_{t}$ are computed by means of a score function (dot product) measuring the similarity between $q_{t}$ and every $k$ in $K$. $z_{t}$ is then obtained by multiplying each attention weight by $v$ in $V$.

\begin{figure}[h!]
    \centering
    \includegraphics[width=0.35\textwidth]{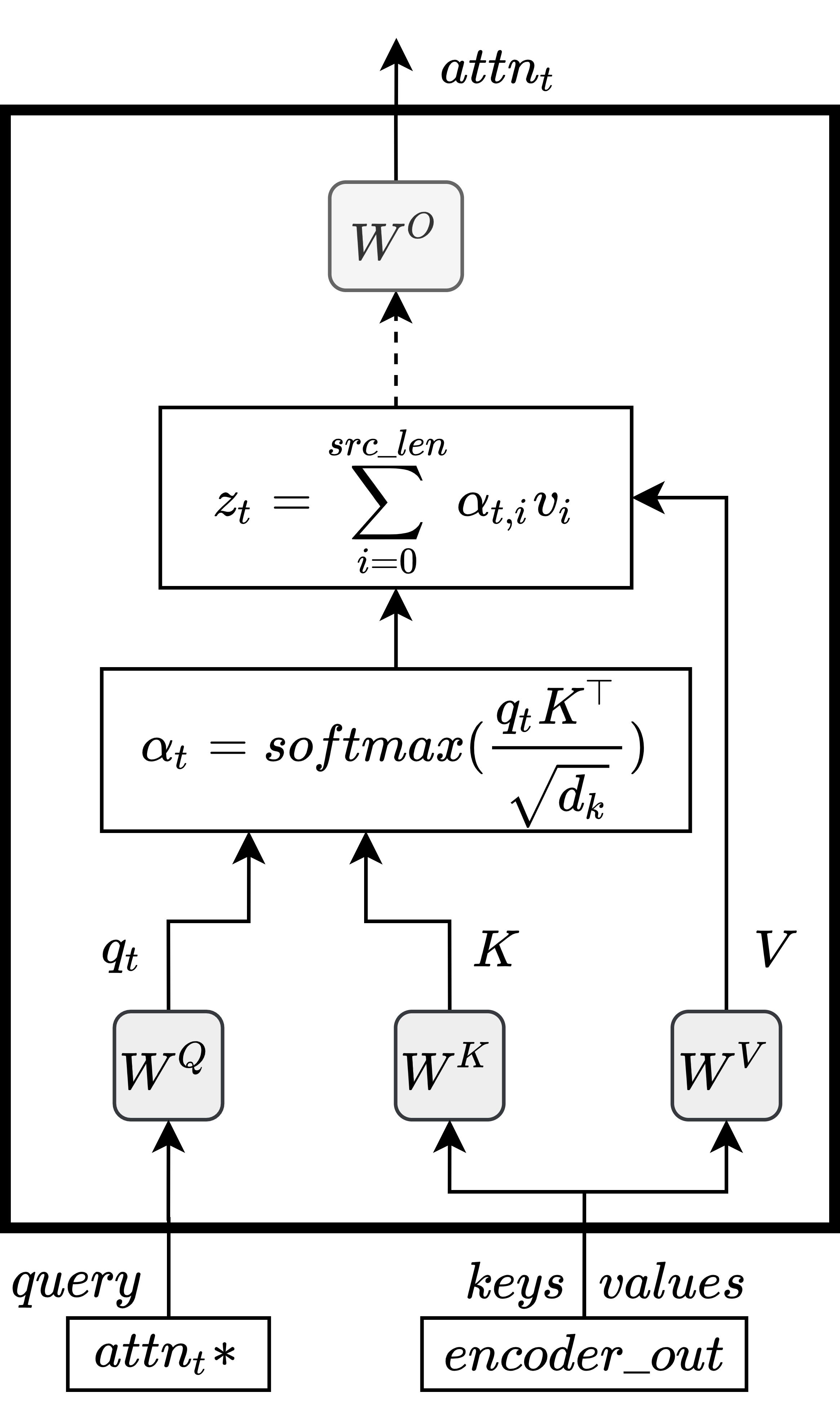}
    \caption{Encoder-decoder attention module of a single head.}
    \label{fig:enc_dec}
\end{figure}

Figure \ref{fig:enc_dec} shows this attention module. Note that, for the sake of simplicity, we only show one head computation.

\begin{figure}[h!]
    \centering
    \includegraphics[width=0.4\textwidth]{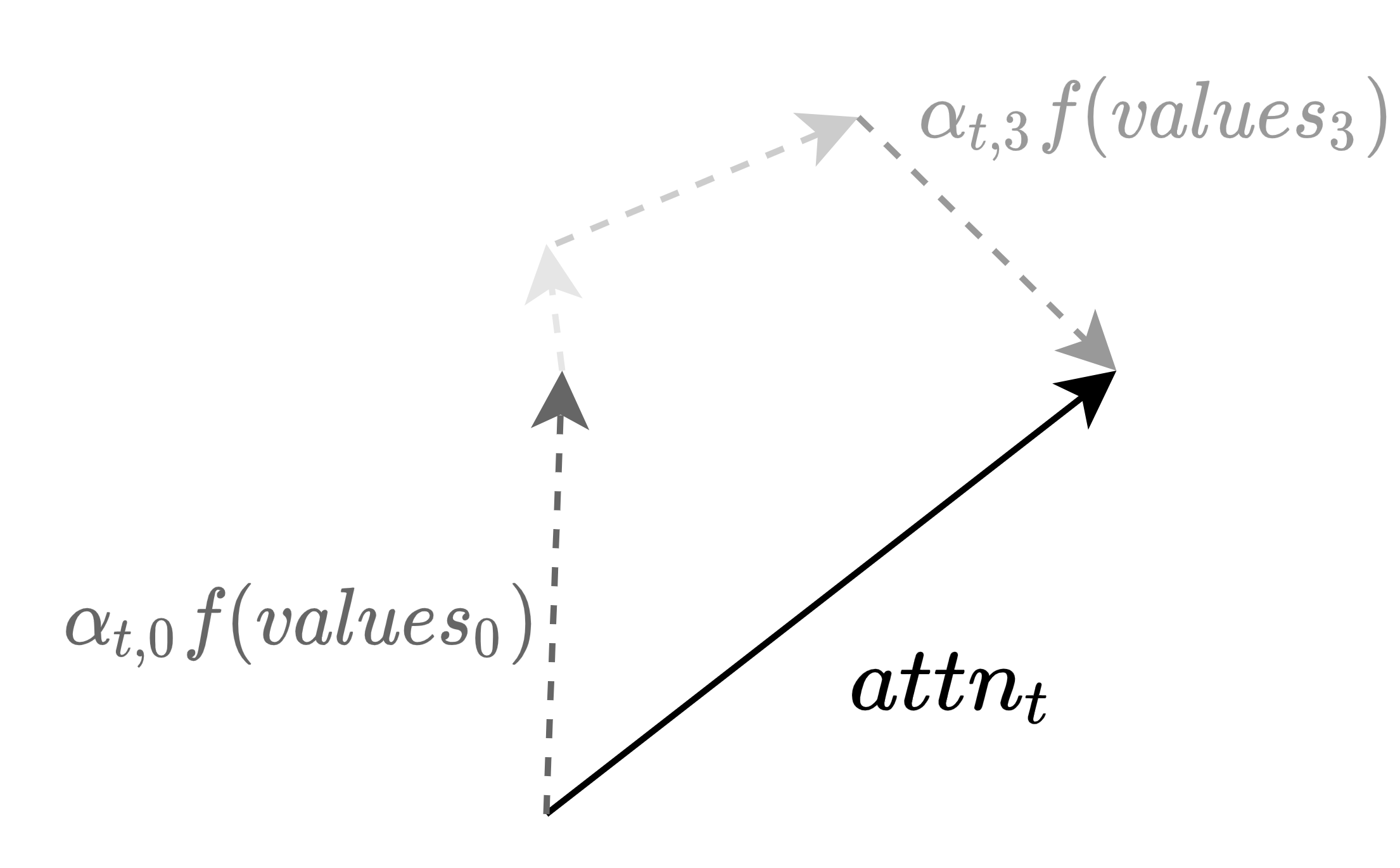}
    \caption{$attn_{t}$ as a vector addition.}
    \label{fig:attn_vector}
\end{figure}

\begin{figure*}[h]
    \centering
    \includegraphics[width=1\textwidth]{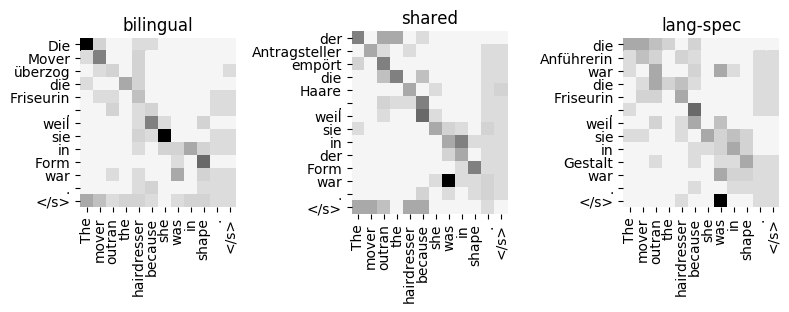}
    \includegraphics[width=1\textwidth]{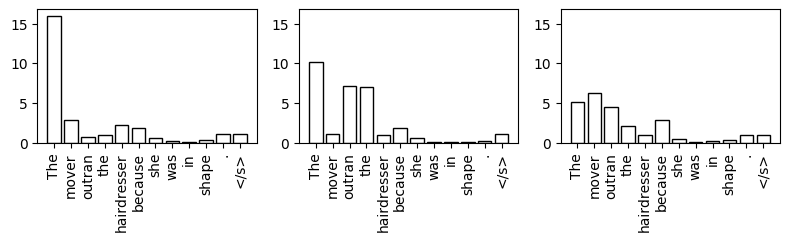}
    \caption{(Top) Attention matrix for Bilingual, Shared and Language-Specific models (layer 2).  Each square corresponds to $\alpha_{t,i}||f(values_i)||$, where $t$ represents the row (target) and $i$ the column (source). (Bottom) Attention distribution, each bar corresponds to $\alpha_{t,i}||f(values_i)||$, where $output_{t}=der/die$. }
    \label{fig:heatmap_1}
\end{figure*}

\begin{figure*}[t!]
    \centering
    \includegraphics[width=1\textwidth]{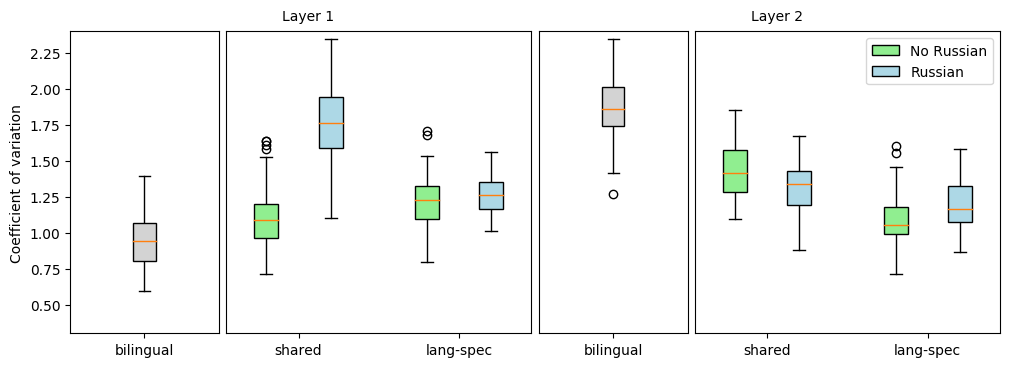}
    \includegraphics[width=1\textwidth]{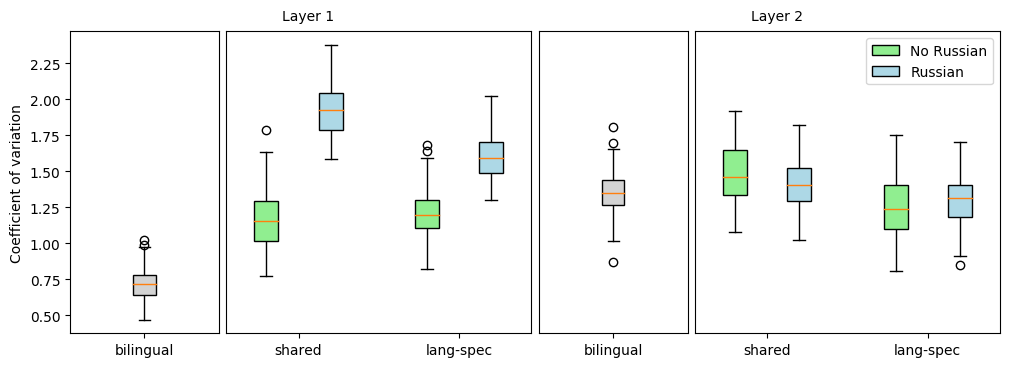}
    \includegraphics[width=1\textwidth]{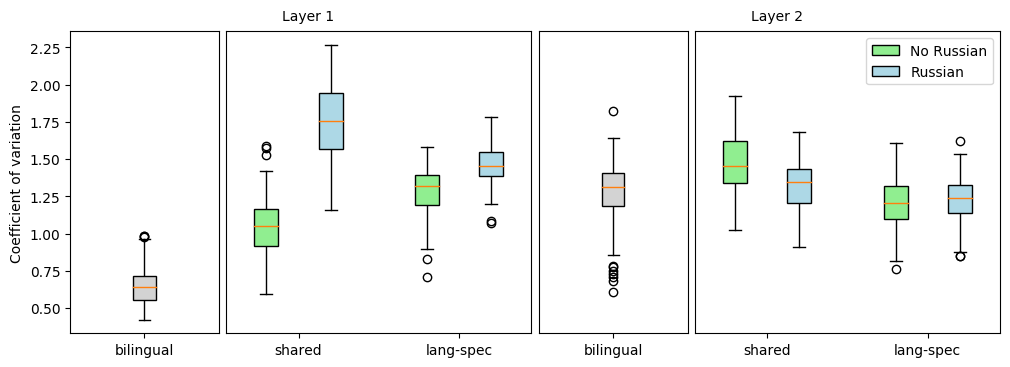}
    \includegraphics[width=1\textwidth]{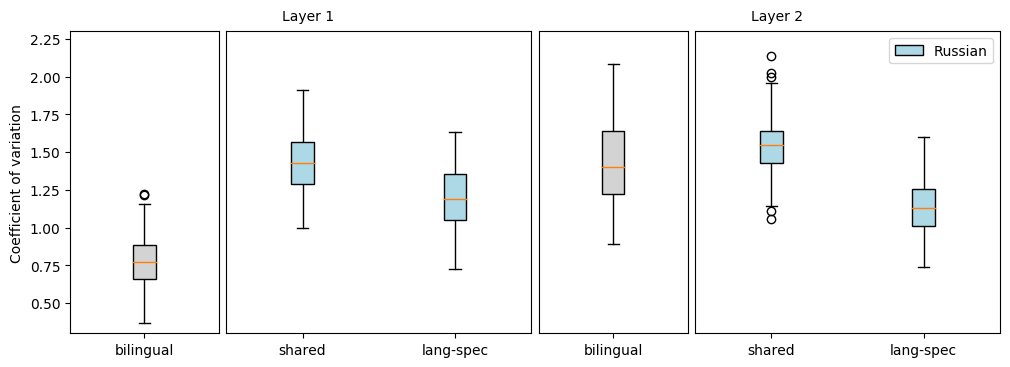}
    \caption{$c_{v}$ when predicting the target determiner for the first 100 sentences in WinoMT: from top to bottom, English-German, English-Spanish, English-French, English-Russian; on the left, Layer 1, on the right, Layer 2}
    \label{fig:cv_comparison}
\end{figure*}

\begin{figure*}[h]
    \centering
    \includegraphics[width=1\textwidth]{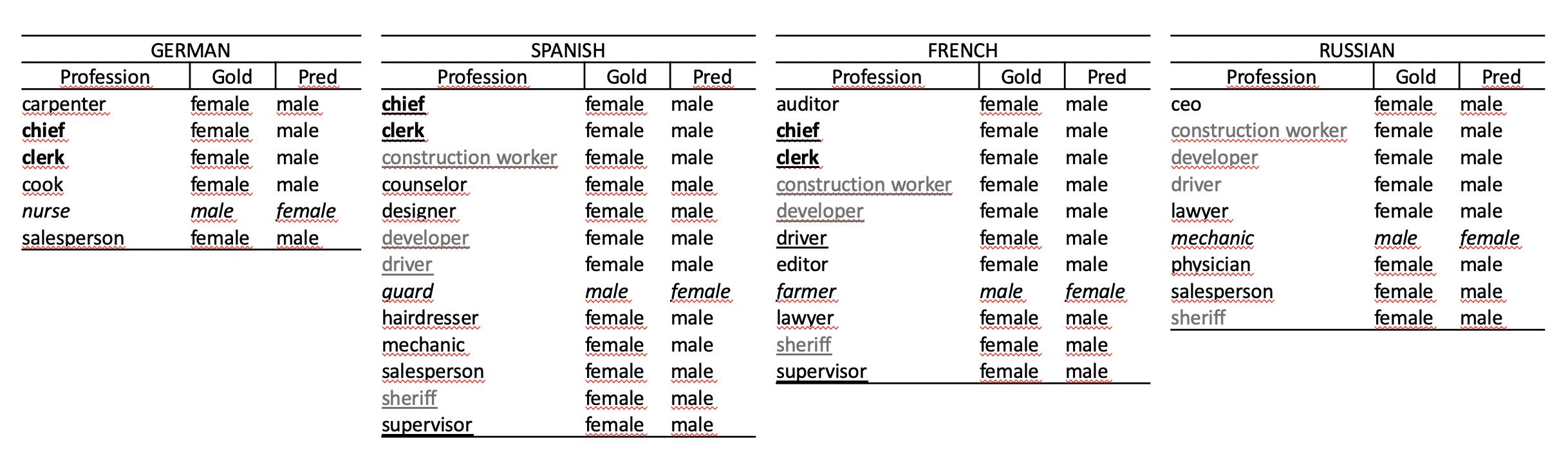}
    \caption{Misclassified occupations in terms of gender. Common words in German, Spanish and French in bold, common words in Spanish and French underlined, common words in Spanish, French and Russian in light grey.}
    \label{fig:occupations}
\end{figure*}



The operations done in the attention module can be reformulated \cite{kobayashi-etal-2020-attention} by moving the linear projection $W^O$ before the $z_{t}$ calculation. In this way, $attn_t$ can be expressed as:
$$
attn_{t} = \left[\sum^{src\_len}_{i=0}\alpha_{t,i} (values_i(W^V + b^V))\right] W^O
$$
Renaming $(values_i(W^V + b^V)) W^O$ as $f(values_i)$ we get:
$$
attn_{t} = \sum^{src\_len}_{i=0}\alpha_{t,i} f(values_i)
$$
Essentially, $attn_{t}$ can be understood as a weighted sum of the transformed vectors. (see Figure \ref{fig:attn_vector}). Then, $||\alpha_{t,i}f(values_{i})||$ is used as a measure of the degree of contribution to $attn_{t}$, hence as the amount of attention given to token at position $i$.

By seeing how much attention is given to each of the input sequence tokens, we can derive conclusions about the decoder's decision-making process.
Attention weights, together with vector norms analysis, have been demonstrated to be a successful technique to measure the degree of contribution of each input token when predicting a target word. Taking the input word with the maximum contribution is used in the word alignment task performing similar to specialized word aligners such as fast\_align \cite{dyer-etal-2013-simple}, and GIZA++ \cite{och03:asc}.

 \paragraph{Coefficient of Variation.} We are interested in the distribution differences of attention between models, i.e. differences between the variability of $\left[\alpha_{t}||f(values)||\right]^{shared}$ and $\left[\alpha_{t}||f(values)||\right]^{lang-spec}$. However, the total amount of attention given to the whole input sequence is different in each model.
   That is: 

\begin{equation*}
\small
[\sum^{src\_len}_{i=0}\alpha_{t,i}||f(values_i)||]^{shared} \neq
\end{equation*}
\begin{equation*}
\small
[\sum^{src\_len}_{i=0}
\alpha_{t,i}||f(values_i)||]^{lang-spec}
\end{equation*}

Therefore, we cannot compare the variances of each distribution directly without previously normalizing to the same scale. Accordingly, we opt to use the coefficient of variation of each model:

\begin{equation*}
c_v=\frac{\sigma(\alpha_{t}||f(values)||)}{\mu(\alpha_{t}||f(values)||)}
\end{equation*}

\paragraph{Bilingual, Shared, Language-Specific Comparison.} Following \cite{kobayashi-etal-2020-attention} results, we focus on the first two decoder layers as the ones that give more reliable information about source-target words relationships (word alignment information). 

For the first layer, Bilingual and Language-Specific have a more uniform attention distribution (see Figure \ref{fig:cv_comparison}), which helps transfer information from the context to extract gender information. The Shared model concentrates its attention much more in all pairs, except for English-Russian, where it approaches the other models. 


For the second layer, the behavior changes depending on the language pair. For English-German, the Bilingual model has more concentrated attention (see Figure \ref{fig:cv_comparison}) followed by the Shared model. In particular, we see an example in Figure \ref{fig:heatmap_1} that shows the differences in the attention spread throughout the input sequence between the different models. We see the attention heatmaps at the top and, at the bottom, the attention distribution when predicting the determiner. There is a higher dependency on the source determiner, especially for the Bilingual model. Considering that for this Bilingual model, the determiner embedding has shown to encode a significant amount of gender information (see Figure \ref{fig:SVM}), concentrating the attention helps predict the translated gender correctly. On the other hand, the Shared does not benefit from this concentration because its determiner's embedding encodes fewer gender information. The Language-Specific continues to be evenly distributed, paying more attention to the determiner and occupation (as shown in the heatmap) and benefiting from the large amount of gender information that these two words encode in their embeddings (see Figure \ref{fig:SVM}).
For the rest of language pairs, the Bilingual model diverts its attention by a big amount, which explains the drop in performance in gender accuracy for this model. For English-Russian, the Language-Specific diverts its attention by a small amount and the Shared model slightly concentrates it. Note that the differences in the coefficient of variation are much smaller for this pair, which is coherent with results in Table \ref{tab:lstmtrans}.


\paragraph{Language Set (without or with Russian).} 
We observe that the coefficient of variation increases when adding Russian both in the Shared and Language-Specific system in the first layer.  
The attention shifts towards the source determiner. This shift is significantly moderated in the Language-Specific case, but it is still beneficial because its source determiners encode gender information (see Figure \ref{fig:SVM}). Whereas, for the Shared system, this shift is much higher and detrimental. It gives less importance to the context, for the benefit of attending its source determiners, the embeddings of which have low gender information. 
This explains why there is a significant variation in gender accuracy in Table \ref{tab:lstmtrans}.

It is essential to notice that the Language-Specific attention behavior is much more robust to adding new languages (especially in the first layer) than the Shared attention.

\section{Manual Analysis}
In this section, we perform a manual analysis of the occupation errors across languages. From previous works  \cite{lewis:2020}, it is demonstrated that culture highly impacts the forms of career-gender terms, where older populations tend to have a stronger association between career and gender. Such impact affects the male/female representations in data \cite{madaan:2018}, where some occupations are represented with the masculine form only, or a higher proportion to males is dedicated.  Our study covers the occupations that are mispredicted 35\% \footnote{This was a trade-off between the percentage of errors and number of sentences enabling us for a manual analysis} of the sentences containing them and at the same time in Bilingual, Shared and Language-Specific systems (the latter two when including Russian). Figure \ref{fig:occupations} shows that mistranslated occupations vary from one language to another. More occupations are mispredicted and translated to have masculine forms than feminine ones, whereas few (in italics) are mistranslated to the feminine form. This conclusion is coherent to previous studies \cite{stanovsky-etal-2019-evaluating}. 
We observe that from the 23 different enlisted words, all of them appear in the top-20 list of errors performed by the Bilingual, Shared and/or Language-Specific (the latter two with Russian) classifiers in section \ref{sec:svm}, see Table \ref{tab:error-list-det} and \ref{tab:error-list-occ} in the appendix. Therefore, the gender information encoded in the source embedding impacts the translation's final gender errors. As follows, we offer a non-exhaustive explanation covering a right proportion of the errors enlisted in Figure \ref{fig:occupations}.

Regarding the German errors, \textit{nurse} tends to be assigned the feminine form (\textit{Krankenschwester = sick + sister}), which is mostly used in everyday language. The masculine form is \textit{Pfleger/Krankenpfleger}, which presents a barely used feminine form \textit{Pflegerin/ Krankenpflegerin}. 

When comparing Romance languages (Spanish and French), standard errors in occupations rise. Because the default gender in Spanish and French was masculine in the past \cite{frank2004gender}, such errors relate to linguistics and culture together. 
In the French culture, masculine forms are predominantly used as gender-neutral, and only
the article may vary for some occupations, like \textit{présidente/président} (CEO), even in the cases where
the feminine form exists. Thus, it is common to say e.g., \textit{madame LE président}, even if the feminine version \textit{madame LA présidente} could be construed as more "politically" correct. 
This can help us explain some of the errors occurring, like the translation of the word (\textit{clerk}), as the \textit{clerk}'s role was dedicated historically to males. Consequently, both languages have only the masculine form, although suitable feminine/masculine translations would be possible. Moreover, some words have the same form for both genders, such as \textit{sheriff}, where only the article differs. 
An interesting example of a feminine mistranslation is the word  \textit{guard}. In Spanish it exists the \textit{guard} (la \textit{guardia}), which has feminine  morphological gender but it refers generally to \textit{police}. 

Regarding the Russian errors, e.g. \textit{driver} (\textit{\foreignlanguage{russian}{водительница}}), although a Russian feminine form exists and understandable for Russian speakers, it is barely used. Other words like \textit{ceo, physician, lawyer} and \textit{sheriff} (\textit{\foreignlanguage{russian}{генеральный директор,доктор, юрист/адвокат}} and \textit{\foreignlanguage{russian}{шериф}}, respectively) do not present a feminine form, being some of them loanwords. 
Finally, the word \textit{mechanic} may be assigned to female because of being wrongly translated and associated with \foreignlanguage{russian}{механика}, which is the field of physics.

\section{Conclusions}

This paper shows that the MNMT architecture by itself has an impact on gender accuracy. Language-Specific outperforms Shared in two different language sets: English, German, Spanish, French and English, German, Spanish, French, and Russian. We observe that the difference in gender accuracy is higher in the language set including Russian. 

Further interpretability analysis of the results shows that source embeddings in the Language-Specific architecture retain higher information on gender. Moreover, this architecture also keeps enough diversion in the attention, especially when including Russian. Both elements help in better inferring the correct gender. 

Finally, a manual analysis shows that most of the errors are made assuming a masculine occupation instead of a feminine one. In contrast, the inverse error tends to come when there is a feminine version of that word with another meaning.  

\section*{Acknowledgements}
The authors want to thank Gemma Boleda, Antonio Calvo, Cristina España, Maite Melero, Júlia Puig, Chantal Winandy and Michel Wolter for their enriching discussions in linguistics biases.

This work is supported by the European Research Council (ERC) under the European Union’s Horizon 2020 research and innovation programme (grant agreement No. 947657).

\bibliography{anthology,custom,naacl}

\begin{thebibliography}{40}
\expandafter\ifx\csname natexlab\endcsname\relax\def\natexlab#1{#1}\fi

\bibitem[{Basta et~al.(2019)Basta, Costa-juss{\`a}, and Casas}]{basta:2019}
Christine Basta, Marta~R Costa-juss{\`a}, and Noe Casas. 2019.
\newblock Evaluating the underlying gender bias in contextualized word
  embeddings.
\newblock \emph{arXiv preprint arXiv:1904.08783}.

\bibitem[{Chen et~al.(2018)Chen, Firat, Bapna, Johnson, Macherey, Foster,
  Jones, Schuster, Shazeer, Parmar, Vaswani, Uszkoreit, Kaiser, Chen, Wu, and
  Hughes}]{chen-etal-2018-best}
Mia~Xu Chen, Orhan Firat, Ankur Bapna, Melvin Johnson, Wolfgang Macherey,
  George Foster, Llion Jones, Mike Schuster, Noam Shazeer, Niki Parmar, Ashish
  Vaswani, Jakob Uszkoreit, Lukasz Kaiser, Zhifeng Chen, Yonghui Wu, and
  Macduff Hughes. 2018.
\newblock \href {https://doi.org/10.18653/v1/P18-1008} {The best of both
  worlds: Combining recent advances in neural machine translation}.
\newblock In \emph{Proceedings of the 56th Annual Meeting of the Association
  for Computational Linguistics (Volume 1: Long Papers)}, pages 76--86,
  Melbourne, Australia.

\bibitem[{Clark et~al.(2019)Clark, Khandelwal, Levy, and
  Manning}]{clark-etal-2019-bert}
Kevin Clark, Urvashi Khandelwal, Omer Levy, and Christopher~D. Manning. 2019.
\newblock \href {https://doi.org/10.18653/v1/W19-4828} {What does {BERT} look
  at? an analysis of {BERT}{'}s attention}.
\newblock In \emph{Proceedings of the 2019 ACL Workshop BlackboxNLP: Analyzing
  and Interpreting Neural Networks for NLP}, pages 276--286, Florence, Italy.

\bibitem[{Cortes and Vapnik(1995)}]{cortes1995support}
Corinna Cortes and Vladimir Vapnik. 1995.
\newblock Support-vector networks.
\newblock \emph{Machine learning}, 20(3):273--297.

\bibitem[{Costa-juss{\`a} et~al.(2020)Costa-juss{\`a}, Li~Lin, and
  Espa{\~n}a-Bonet}]{costa-jussa-etal-2020-gebiotoolkit}
Marta~R. Costa-juss{\`a}, Pau Li~Lin, and Cristina Espa{\~n}a-Bonet. 2020.
\newblock \href {https://www.aclweb.org/anthology/2020.lrec-1.502}
  {{G}e{B}io{T}oolkit: Automatic extraction of gender-balanced multilingual
  corpus of {W}ikipedia biographies}.
\newblock In \emph{Proceedings of The 12th Language Resources and Evaluation
  Conference}, pages 4081--4088, Marseille, France.

\bibitem[{Costa-jussà(2019)}]{costajussa:2019}
Marta~R. Costa-jussà. 2019.
\newblock An analysis of gender bias studies in natural language processing.
\newblock \emph{Nature Machine Intelligence}, 1(11):495--496.

\bibitem[{Costa-jussà and de~Jorge(2020)}]{costajussa:2020}
Marta~R. Costa-jussà and Adrià de~Jorge. 2020.
\newblock Fine-tunning neural machine translation on gender-balanced datasets.
\newblock In \emph{Scond Workshop on Gender Bias in Natural Language
  Processing}.

\bibitem[{Dyer et~al.(2013)Dyer, Chahuneau, and Smith}]{dyer-etal-2013-simple}
Chris Dyer, Victor Chahuneau, and Noah~A. Smith. 2013.
\newblock \href {https://www.aclweb.org/anthology/N13-1073} {A simple, fast,
  and effective reparameterization of {IBM} model 2}.
\newblock In \emph{Proceedings of the 2013 Conference of the North {A}merican
  Chapter of the Association for Computational Linguistics: Human Language
  Technologies}, pages 644--648, Atlanta, Georgia.

\bibitem[{Escolano et~al.(2020)Escolano, Costa-juss{\`a}, Fonollosa, and
  Artetxe}]{escolano:2020}
Carlos Escolano, Marta~R. Costa-juss{\`a}, Jos{\'e} A.~R. Fonollosa, and
  M\'ikel Artetxe. 2020.
\newblock Multilingual machine translation: Closing the gap between shared and
  language-specific encoder-decoders.
\newblock \emph{Corr}, abs/2004.06575.

\bibitem[{Firat et~al.(2017)Firat, Cho, Sankaran, Vural, and
  Bengio}]{firat:2017}
Orhan Firat, Kyunghyun Cho, Baskaran Sankaran, Fatos T.~Yarman Vural, and
  Yoshua Bengio. 2017.
\newblock {Multi-Way, Multilingual Neural Machine Translation}.
\newblock \emph{Computer Speech and Language, Special Issue in Deep learning
  for Machine Translation}.

\bibitem[{Font and Costa-juss{\`a}(2019)}]{font:2019}
Joel~Escud{\'e} Font and Marta~R. Costa-juss{\`a}. 2019.
\newblock \href {https://doi.org/10.18653/v1/W19-3821} {Equalizing gender bias
  in neural machine translation with word embeddings techniques}.
\newblock In \emph{Proceedings of the First ACL Workshop on Gender Bias in
  Natural Language Processing}, pages 147--154, Florence, Italy.

\bibitem[{Frank et~al.(2004)Frank, Hoffmann, Strobel et~al.}]{frank2004gender}
Anke Frank, Chr Hoffmann, Maria Strobel, et~al. 2004.
\newblock Gender issues in machine translation.
\newblock \emph{Univ. Bremen}.

\bibitem[{Gebru et~al.(2018)Gebru, Morgenstern, Vecchione, Vaughan, Wallach,
  III, and Crawford}]{gebru}
Timnit Gebru, Jamie Morgenstern, Briana Vecchione, Jennifer~Wortman Vaughan,
  Hanna~M. Wallach, Hal~Daum{\'{e}} III, and Kate Crawford. 2018.
\newblock \href {http://arxiv.org/abs/1803.09010} {Datasheets for datasets}.
\newblock \emph{CoRR}, abs/1803.09010.

\bibitem[{Htut et~al.(2019)Htut, Phang, Bordia, and Bowman}]{Htut2019DoAH}
Phu~Mon Htut, Jason Phang, Shikha Bordia, and Samuel~R. Bowman. 2019.
\newblock Do attention heads in bert track syntactic dependencies?
\newblock \emph{ArXiv}, abs/1911.12246.

\bibitem[{Johnson et~al.(2017)Johnson, Schuster, Le, Krikun, Wu, Chen, Thorat,
  Vi{\'e}gas, Wattenberg, Corrado et~al.}]{johnson2017google}
Melvin Johnson, Mike Schuster, Quoc~V Le, Maxim Krikun, Yonghui Wu, Zhifeng
  Chen, Nikhil Thorat, Fernanda Vi{\'e}gas, Martin Wattenberg, Greg Corrado,
  et~al. 2017.
\newblock Google’s multilingual neural machine translation system: Enabling
  zero-shot translation.
\newblock \emph{Transactions of the Association for Computational Linguistics},
  5:339--351.

\bibitem[{Kingma and Ba(2014)}]{kingma2014adam}
Diederik~P Kingma and Jimmy Ba. 2014.
\newblock Adam: A method for stochastic optimization.
\newblock \emph{arXiv preprint arXiv:1412.6980}.

\bibitem[{Kobayashi et~al.(2020)Kobayashi, Kuribayashi, Yokoi, and
  Inui}]{kobayashi-etal-2020-attention}
Goro Kobayashi, Tatsuki Kuribayashi, Sho Yokoi, and Kentaro Inui. 2020.
\newblock \href {https://doi.org/10.18653/v1/2020.emnlp-main.574} {Attention is
  not only a weight: Analyzing transformers with vector norms}.
\newblock In \emph{Proceedings of the 2020 Conference on Empirical Methods in
  Natural Language Processing (EMNLP)}, pages 7057--7075, Online.

\bibitem[{Koehn(2005)}]{koehn:2005}
Philipp Koehn. 2005.
\newblock Europarl: A parallel corpus for statistical machine translation.
\newblock In \emph{MT summit}, volume~5, pages 79--86. Citeseer.

\bibitem[{Koehn et~al.(2007)Koehn, Hoang, Birch, Callison-Burch, Federico,
  Bertoldi, Cowan, Shen, Moran, Zens et~al.}]{koehn2007moses}
Philipp Koehn, Hieu Hoang, Alexandra Birch, Chris Callison-Burch, Marcello
  Federico, Nicola Bertoldi, Brooke Cowan, Wade Shen, Christine Moran, Richard
  Zens, et~al. 2007.
\newblock Moses: Open source toolkit for statistical machine translation.
\newblock In \emph{Proceedings of the ACL: Demo Papers}, pages 177--180.

\bibitem[{Kovaleva et~al.(2019)Kovaleva, Romanov, Rogers, and
  Rumshisky}]{kovaleva-etal-2019-revealing}
Olga Kovaleva, Alexey Romanov, Anna Rogers, and Anna Rumshisky. 2019.
\newblock \href {https://doi.org/10.18653/v1/D19-1445} {Revealing the dark
  secrets of {BERT}}.
\newblock In \emph{Proceedings of the 2019 Conference on Empirical Methods in
  Natural Language Processing and the 9th International Joint Conference on
  Natural Language Processing (EMNLP-IJCNLP)}, pages 4365--4374, Hong Kong,
  China.

\bibitem[{Kuczmarski and Johnson(2018)}]{johnson}
James Kuczmarski and Melvin Johnson. 2018.
\newblock Gender-aware natural language translation.

\bibitem[{Lewis and Lupyan(2020)}]{lewis:2020}
Moly Lewis and Gary Lupyan. 2020.
\newblock Gender stereotypes are reflected in the distributional structure of
  25 languages.
\newblock \emph{Nat Hum Behav 4}, page 1021–1028.

\bibitem[{Lin et~al.(2019)Lin, Tan, and Frank}]{lin-etal-2019-open}
Yongjie Lin, Yi~Chern Tan, and Robert Frank. 2019.
\newblock \href {https://doi.org/10.18653/v1/W19-4825} {Open sesame: Getting
  inside {BERT}{'}s linguistic knowledge}.
\newblock In \emph{Proceedings of the 2019 ACL Workshop BlackboxNLP: Analyzing
  and Interpreting Neural Networks for NLP}, pages 241--253, Florence, Italy.

\bibitem[{Lu et~al.(2018)Lu, Keung, Ladhak, Bhardwaj, Zhang, and
  Sun}]{lu-etal-2018-neural}
Yichao Lu, Phillip Keung, Faisal Ladhak, Vikas Bhardwaj, Shaonan Zhang, and
  Jason Sun. 2018.
\newblock \href {https://doi.org/10.18653/v1/W18-6309} {A neural interlingua
  for multilingual machine translation}.
\newblock In \emph{Proceedings of the Third Conference on Machine Translation:
  Research Papers}, pages 84--92, Brussels, Belgium.

\bibitem[{Madaan et~al.(2018{\natexlab{a}})Madaan, Mehta, Agrawaal, Malhotra,
  Aggarwal, and Saxena}]{bollywood}
Nishtha Madaan, Sameep Mehta, Taneea~S. Agrawaal, Vrinda Malhotra, Aditi
  Aggarwal, and Mayank Saxena. 2018{\natexlab{a}}.
\newblock \href {http://arxiv.org/abs/1710.04117} {Analyzing gender
  stereotyping in bollywood movies}.
\newblock In \emph{Proceedings of Machine Learning Research 81:1:14}.

\bibitem[{Madaan et~al.(2018{\natexlab{b}})Madaan, Singh, Mehta, Chetan, and
  Joshi}]{madaan:2018}
Nishtha Madaan, Gautam Singh, Sameep Mehta, Aditya Chetan, and Brihi Joshi.
  2018{\natexlab{b}}.
\newblock Generating clues for gender based occupation de-biasing in text.
\newblock \emph{arXiv preprint arXiv:1804.03839}.

\bibitem[{Mare{\v{c}}ek and Rosa(2019)}]{marecek-rosa-2019-balustrades}
David Mare{\v{c}}ek and Rudolf Rosa. 2019.
\newblock \href {https://doi.org/10.18653/v1/W19-4827} {From balustrades to
  pierre vinken: Looking for syntax in transformer self-attentions}.
\newblock In \emph{Proceedings of the 2019 ACL Workshop BlackboxNLP: Analyzing
  and Interpreting Neural Networks for NLP}, pages 263--275, Florence, Italy.

\bibitem[{Och and Ney(2003)}]{och03:asc}
Franz~Josef Och and Hermann Ney. 2003.
\newblock A systematic comparison of various statistical alignment models.
\newblock \emph{Computational Linguistics}, 29(1):19--51.

\bibitem[{Prates et~al.(2020)Prates, Avelar, and Lamb}]{prates:2018}
Marcelo O.~R. Prates, Pedro H.~C. Avelar, and Luis Lamb. 2020.
\newblock Assessing gender bias in machine translation -- a case study with
  google translate.
\newblock \emph{Neural Computing and Applications, 32}, page 6363–6381.

\bibitem[{Raganato and Tiedemann(2018)}]{raganato-tiedemann-2018-analysis}
Alessandro Raganato and J{\"o}rg Tiedemann. 2018.
\newblock \href {https://doi.org/10.18653/v1/W18-5431} {An analysis of encoder
  representations in transformer-based machine translation}.
\newblock In \emph{Proceedings of the 2018 {EMNLP} Workshop {B}lackbox{NLP}:
  Analyzing and Interpreting Neural Networks for {NLP}}, pages 287--297,
  Brussels, Belgium.

\bibitem[{Reif et~al.(2019)Reif, Yuan, Wattenberg, Viegas, Coenen, Pearce, and
  Kim}]{NEURIPS2019_159c1ffe}
Emily Reif, Ann Yuan, Martin Wattenberg, Fernanda~B Viegas, Andy Coenen, Adam
  Pearce, and Been Kim. 2019.
\newblock \href
  {https://proceedings.neurips.cc/paper/2019/file/159c1ffe5b61b41b3c4d8f4c2150f6c4-Paper.pdf}
  {Visualizing and measuring the geometry of bert}.
\newblock In \emph{Advances in Neural Information Processing Systems},
  volume~32, pages 8594--8603. Curran Associates, Inc.

\bibitem[{Rudinger et~al.(2018)Rudinger, Naradowsky, Leonard, and
  Van~Durme}]{rudinger:2018}
Rachel Rudinger, Jason Naradowsky, Brian Leonard, and Benjamin Van~Durme. 2018.
\newblock Gender bias in coreference resolution.
\newblock \emph{arXiv preprint arXiv:1804.09301}.

\bibitem[{Sennrich et~al.(2016)Sennrich, Haddow, and
  Birch}]{sennrich-etal-2016-neural}
Rico Sennrich, Barry Haddow, and Alexandra Birch. 2016.
\newblock \href {https://doi.org/10.18653/v1/P16-1162} {Neural machine
  translation of rare words with subword units}.
\newblock In \emph{Proceedings of the 54th Annual Meeting of the Association
  for Computational Linguistics (Volume 1: Long Papers)}, pages 1715--1725,
  Berlin, Germany.

\bibitem[{Stanovsky et~al.(2019)Stanovsky, Smith, and
  Zettlemoyer}]{stanovsky-etal-2019-evaluating}
Gabriel Stanovsky, Noah~A. Smith, and Luke Zettlemoyer. 2019.
\newblock \href {https://doi.org/10.18653/v1/P19-1164} {Evaluating gender bias
  in machine translation}.
\newblock In \emph{Proceedings of the 57th Annual Meeting of the Association
  for Computational Linguistics}, pages 1679--1684, Florence, Italy.

\bibitem[{Sutskever et~al.(2014)Sutskever, Vinyals, and Le}]{Sutskever:2014}
Ilya Sutskever, Oriol Vinyals, and Quoc~V. Le. 2014.
\newblock Sequence to sequence learning with neural networks.
\newblock In \emph{Annual Conference on Neural Information Processing Systems},
  pages 3104--3112.

\bibitem[{Tang et~al.(2018)Tang, Sennrich, and Nivre}]{tang-etal-2018-analysis}
Gongbo Tang, Rico Sennrich, and Joakim Nivre. 2018.
\newblock \href {https://doi.org/10.18653/v1/W18-6304} {An analysis of
  attention mechanisms: The case of word sense disambiguation in neural machine
  translation}.
\newblock In \emph{Proceedings of the Third Conference on Machine Translation:
  Research Papers}, pages 26--35, Brussels, Belgium.

\bibitem[{Vaswani et~al.(2017)Vaswani, Shazeer, Parmar, Uszkoreit, Jones,
  Gomez, Kaiser, and Polosukhin}]{vaswani_attention:2017}
Ashish Vaswani, Noam Shazeer, Niki Parmar, Jakob Uszkoreit, Llion Jones,
  Aidan~N Gomez, {\L}ukasz Kaiser, and Illia Polosukhin. 2017.
\newblock Attention is all you need.
\newblock In \emph{Advances in neural information processing systems}, pages
  5998--6008.

\bibitem[{Webster et~al.(2018)Webster, Recasens, Axelrod, and
  Baldridge}]{kelliegap:2018}
Kellie Webster, Marta Recasens, Vera Axelrod, and Jason Baldridge. 2018.
\newblock Mind the {GAP:} {A} balanced corpus of gendered ambiguous pronouns.
\newblock \emph{CoRR}, abs/1810.05201.

\bibitem[{Webster et~al.(2020)Webster, Wang, Tenney, Beutel, Pitler, Pavlick,
  Chen, and Petrov}]{webster2020measuring}
Kellie Webster, Xuezhi Wang, Ian Tenney, Alex Beutel, Emily Pitler, Ellie
  Pavlick, Jilin Chen, and Slav Petrov. 2020.
\newblock \href {http://arxiv.org/abs/2010.06032} {Measuring and reducing
  gendered correlations in pre-trained models}.

\bibitem[{Zmigrod et~al.(2019)Zmigrod, Mielke, Wallach, and
  Cotterell}]{zmigrod-etal-2019-counterfactual}
Ran Zmigrod, Sabrina~J. Mielke, Hanna Wallach, and Ryan Cotterell. 2019.
\newblock \href {https://doi.org/10.18653/v1/P19-1161} {Counterfactual data
  augmentation for mitigating gender stereotypes in languages with rich
  morphology}.
\newblock In \emph{Proceedings of the 57th Annual Meeting of the Association
  for Computational Linguistics}, pages 1651--1661, Florence, Italy.

\end{thebibliography}
\bibliographystyle{acl_natbib}

\appendix

\section{Classification Error List}
\label{sec:appendix-cls-list}

Table \ref{tab:error-list-det} and \ref{tab:error-list-occ} report the top-20 error word list from classification models trained in section \ref{sec:svm}.

\begin{table*}[]
\small
\begin{tabular}{|l|l|l|l|l|l|l|l|l|}
\hline
\textbf{\#} & \textbf{Bil\_ende} & \textbf{Bil\_enes} & \textbf{Bil\_enfr} & \textbf{Bil\_enru} & \textbf{Shared} & \textbf{Shared\_ru} & \textbf{LangSpec} & \textbf{LangSpec\_ru} \\ \hline
1           & carpenter          & someone            & someone            & someone            & someone         & someone             & carpenter         & someone               \\
2           & cleaner            & physician          & carpenter          & janitor            & janitor         & janitor             & someone           & nurse                 \\
3           & librarian          & janitor            & janitor            & teacher            & teacher         & teacher             & hairdresser       & sheriff               \\
4           & CEO                & carpenter          & teacher            & carpenter          & carpenter       & carpenter           & developer         & cashier               \\
5           & mechanic           & manager            & manager            & manager            & manager         & manager             & librarian         & carpenter             \\
6           & someone            & librarian          & physician          & physician          & physician       & physician           & cleaner           & guard                 \\
7           & cook               & nurse              & supervisor         & nurse              & nurse           & nurse               & cashier           & Someone               \\
8           & baker              & accountant         & nurse              & accountant         & cashier         & librarian           & recepcionist      & cleaner               \\
9           & clerk              & teacher            & librarian          & librarian          & librarian       & accountant          & nurse             & cook                  \\
10          & Someone            & clerk              & receptionist       & developer          & accountant      & supervisor          & janitor           & librarian             \\
11          & guard              & cleaner            & salesperson        & cleaner            & developer       & developer           & sheriff           & hairdresser           \\
12          & housekeeper        & cashier            & hairdresser        & cashier            & cleaner         & cleaner             & tailor            & receptionist          \\
13          & tailor             & receptionist       & tailor             & receptionist       & receptionist    & cashier             & mechanic          & chief                 \\
14          & janitor            & writer             & baker              & writer             & writer          & receptionist        & baker             & clerk                 \\
15          & customer           & mechanic           & developer          & hairdresser        & hairdresser     & writer              & housekeeper       & customer              \\
16          & nurse              & baker              & cleaner            & supervisor         & supervisor      & mechanic            & physician         & child                 \\
17          & patient            & chief              & cashier            & mechanic           & mechanic        & clerk               & CEO               & baker                 \\
18          & child              & hairdresser        & writer             & baker              & clerk           & baker               & counselor         & tailor                \\
19          & physician          & CEO                & counselor          & chief              & baker           & chief               & construction      & laborer               \\
20          & manager            & supervisor         & mechanic           & lawyer             & chief           & hairdresser         & salesperson       & counselor             \\ \hline
\end{tabular}
\caption{List of the 20 most common misclassified accupations by the SVM models trained with determiners}
\label{tab:error-list-det}
\end{table*}
\vspace{-5cm}

\begin{table*}[h!]
\small
\begin{tabular}{|l|l|l|l|l|l|l|l|l|}
\hline
\textbf{\#} & \textbf{Bil\_ende} & \textbf{Bil\_enes} & \textbf{Bil\_enfr} & \textbf{Bil\_enru} & \textbf{Shared} & \textbf{Shared\_ru} & \textbf{LangSpec} & \textbf{LangSpec\_ru} \\ \hline
1           & someone            & someone            & someone            & someone            & someone         & someone             & someone           & someone               \\
2           & CEO                & housekeeper        & manager            & teacher            & cashier         & nurse               & accountant        & nurse                 \\
3           & carpenter          & nurse              & teacher            & librarian          & nurse           & cashier             & hairdresser       & sheriff               \\
4           & clerk              & writer             & physician          & mechanic           & librarian       & mechanic            & receptionist      & cashier               \\
5           & cook               & cashier            & carpenter          & accountant         & attendant       & cleaner             & clerk             & carpenter             \\
6           & cleaner            & cleaner            & baker              & physician          & cook            & attendant           & designer          & guard                 \\
7           & librarian          & clerk              & nurse              & cleaner            & clerk           & guard               & lawyer            & Someone               \\
8           & nurse              & manager            & librarian          & writer             & guard           & librarian           & driver            & cleaner               \\
9           & guard              & chief              & mechanic           & cashier            & janitor         & receptionist        & physician         & cook                  \\
10          & Someone            & counselor          & hairdresser        & carpenter          & carpenter       & cook                & librarian         & librarian             \\
11          & chief              & mechanic           & clerk              & clerk              & Someone         & Someone             & baker             & hairdresser           \\
12          & tailor             & librarian          & tailor             & receptionist       & laborer         & janitor             & auditor           & receptionist          \\
13          & mechanic           & CEO                & mover              & lawyer             & mover           & baker               & writer            & chief                 \\
14          & physician          & carpenter          & chief              & baker              & baker           & CEO                 & supervisor        & clerk                 \\
15          & baker              & developer          & attendant          & auditor            & designer        & salesperson         & nurse             & customer              \\
16          & customer           & janitor            & counselor          & chief              & mechanic        & tailor              & cleaner           & child                 \\
17          & janitor            & teacher            & salesperson        & supervisor         & cleaner         & mover               & analyst           & baker                 \\
18          & child              & editor             & auditor            & CEO                & tailor          & clerk               & teacher           & tailor                \\
19          & supervisor         & auditor            & cashier            & hairdresser        & salesperson     & hairdresser         & farmer            & laborer               \\
20          & counselor          & Someone            & housekeeper        & nurse              & manager         & carpenter           & cook              & counselor             \\ \hline
\end{tabular}
\caption{List of the 20 most common misclassified accupations by the SVM models trained with occupations.}
\label{tab:error-list-occ}
\end{table*}


\end{document}